\documentclass[runningheads]{llncs}
\usepackage{graphicx}
\graphicspath{ {./images/} }

\usepackage{algorithm} 
\usepackage{algpseudocode} 

\algdef{SE}[DOWHILE]{Do}{doWhile}{\algorithmicdo}[1]{\algorithmicwhile\ #1}%

\usepackage{amsmath}

\usepackage[toc,page]{appendix}

\usepackage[
backend=biber,
style=numeric,
citestyle=numeric,
]{biblatex}

\usepackage{epigraph}

\begin{document}
\title{Multi-level Data Representation For Training Deep Helmholtz Machines}

%
%
\author{Jose Miguel Ramos jose.miguel.ramos@tecnico.ulisboa.pt \\ Luis Sa-Couto luis.sa.couto@tecnico.ulisboa.pt \\ Andreas Wichert andreas.wichert@tecnico.ulisboa.pt}
\authorrunning{J. M. Ramos, L. Sa-Couto, A. Wichert}
%
\institute{Department of Computer Science and Engineering, INESC-ID \& Instituto Superior Técnico, University of Lisbon, 2744-016 Porto Salvo, Portugal}
\maketitle 

\begin{abstract}

A vast majority of the current research in the field of Machine Learning is done using algorithms with strong arguments pointing to their biological implausibility such as Backpropagation, deviating the field's focus from understanding its original organic inspiration to a compulsive search for optimal performance.
Yet, there have been a few proposed models that respect most of the biological constraints present in the human brain and are valid candidates for mimicking some of its properties and mechanisms.
In this paper, we will focus on guiding the learning of a biologically plausible generative model called the Helmholtz Machine in complex search spaces using a heuristic based on the Human Image Perception mechanism. 
We hypothesize that this model's learning algorithm is not fit for Deep Networks due to its Hebbian-like local update rule, rendering it incapable of taking full advantage of the compositional properties that multi-layer networks provide.
We propose to overcome this problem, by providing the network's hidden layers with visual queues at different resolutions using a Multi-level Data representation.
The results on several image datasets showed the model was able to not only obtain better overall quality but also a wider diversity in the generated images, corroborating our intuition that using our proposed heuristic allows the model to take more advantage of the network's depth growth. More importantly, they show the unexplored possibilities underlying brain-inspired models and techniques.

\end{abstract}

\section{Introduction}

Most recent machine learning models have shown great effectiveness at solving a wide range of complex cognitive tasks \cite{whittington2019,lecun2015}, and back-propagation algorithms seem to be at the core of the majority of those models, proving it to be one of the most reliable and fast ways for machines to learn \cite{bartunov2018, marblestone2016, lillicrap2020}. Visual pattern recognition is one of the many fields in which back-propagation algorithms thrive \cite{sutskever2013, lecun2015, goodfellow2016}. The evolution of these models' quality has been impressively swift, but as we get closer to perfection, the possible improvements get evermore difficult \cite{ahlawat2020}. For some of the more simple visual tasks like image classification of handwritten digits in the famous MNIST dataset \cite{lecun1998}, these models have surpassed the brain capabilities, performing better than human participants \cite{ahlawat2020, ciregan2012}.

The surpassing of the human brain's accuracy is an amazing scientific mark and allows for more reliable and robust technology.

In the midst of this search for better and more powerful models, grew a firmer and firmer connection between the two concepts of intelligence and accuracy.
We seem to have been intuitively led to the conclusion that the better a model performs at a certain task, the more intelligent it is. In a sense, we deviate from trying to mimic the brain's biological way of processing information and focus instead on neural network models that perform better \cite{marblestone2016, krotov2019}.

Nonetheless, even if there are models that compete with the human brain at performing specific tasks, there is no model that comes close to the robustness and flexibility of the human brain when dealing with general image classification and pattern recognition problems.

Therefore, a large part of the scientific community is still focused on the biologically plausible side of machine learning, proposing new competitive models that remain an arguably plausible implementation of some human brain mechanisms and properties \cite{krotov2019, illing2019, bengio2015, ravichandran2020, sa2019}.

\subsection{Back-propagation's Biological Plausibility}

Despite the obvious biological inspiration of the Back-propagation (Backprop) algorithm \cite{marblestone2016, mcculloch1943}, its biological plausibility has been questioned very early on from its appearance \cite{crick1989, stork1989}. In recent years, although there have been many attempts to create biologically plausible and empirically powerful learning algorithms similar to Backprop \cite{lillicrap2016, bartunov2018, krotov2019}, there is an overall consensus that some fundamental properties of back-propagation are too difficult for the human brain to implement \cite{illing2019, marblestone2016}. 

The first and most relevant argument is related to the fact that backprop synaptic weight updates depend on computations and activation on an entire chain of neurons whereas biological synapses change their connection strength solely based on local signals. Furthermore, for this Gradient-based algorithm to work, biological neurons' updates would have to be frozen in time waiting for the signal to reach its final destination where the error comparison is made, and only after the signal travels backwards the membrane permeability would be changed in accordance to its success or failure \cite{bengio2015}.

The second is the fact that back-propagation uses the same weights when performing forward and backwards passes, which would require identical bidirectional connections in biological neurons that are not present in all parts of the brain. 

And lastly, the fact backprop networks propagate firing probabilities, whereas biological neurons only propagate neuron spikes \cite{whittington2019}.

\subsection{Helmholtz Machines' Biological Inspiration}

We propose to look at an older Generative model called Helmholtz Machine (HM) \cite{dayan1995}, which uses the Wake-Sleep (WS) algorithm \cite{hinton1995} (details in Appendix \ref{WS_details}) instead of Back-propagation.

The Wake-Sleep is an unsupervised learning algorithm that uses two different networks to simultaneously learn a predictive Recognition Model and a generative Generation Model. Despite not being a completely Hebbian algorithm, its activation and learning rules are as local as the Hebb rule \cite{neal1997}.

Hebbian learning algorithms respect the original proposition made by Hebb \cite{hebb1949}, that learning and memory in the brain would arise from increased synaptic efficacy, triggered by the coordinated firing of the pre- and post-synaptic neurons \cite{sumner2020}, and more importantly, they solve the previously mentioned locality problem because the synaptic weight updates only depend on the previous layer. Thus, the locality of WS also helps to avoid that problem in a similar way to the Hebbian rule. 

The unsupervised nature of the algorithm, also contributes to its plausibility, since the human brain's learning is mostly done with unsupervised data. And unlike in Back-propagation where it is very difficult to find an implementation that works by propagating neuron activations instead of firing probabilities, the WS algorithm can work effectively with both options, solving the third mentioned back-propagation implausibility argument.

Furthermore, the learning algorithm of these machines is based on the biological idea of being awake and asleep. Its intuition is that after we experience an event, we also produce our own variations of those events. 
This idea can be easily extrapolated to what happens on a big scale daily, where we experience reality during our wake phase, and then recreate it in our sleep, but there is a shorter scale example that perhaps compares better to the actual behavior of the model that occurs, for example, in the interaction between the human eyes and the brain. Our brain receives continuous streams of images that our eyes are capturing, and while we are receiving them, we subconsciously try to predict what will happen in the next frame, and when the reality does not match your expectation, for example, when a magician pulls a rabbit out of the hat, we become surprised.
The HM network also mimics this behavior, and after receiving an observation from the world, it will produce a dream, then the network will adjust its weights in order to create more plausible dreams, and try to reduce the surprise when experiencing the next event. Likewise, if you see the same magic trick performed enough times, you will learn to expect what was previously unexpected.

This ``reduce of surprise" corresponds to minimizing a quantity very imminent in neuro-scientific research called Free Energy \cite{friston2005,friston2006}, which is ``an information theory measure that bounds the surprise on sampling some data, given a generative model" \cite{friston2009}. Thus, the minimization of Free Energy corroborates the hypothesis that ``a biological agent resists the tendency toward disorder through a minimization of uncertainty" \cite{sumner2020, friston2009, friston2012} alluded to in the previous example.

\section{Improving Wake-Sleep}

In spite of the WS algorithm being interesting from a neuro-scientific perspective, its' lack of efficiency \cite{kingma2019} and ability to perform as well as other learning algorithms have led it to be less and less explored in recent years. One of its biggest disadvantages is that when the complexity of the network increases, the algorithm's performance starts to be less impressive. If the complexity of the world we are trying to mimic increases, our model needs to be able to capture higher-level abstractions and generalize better, which can be done by increasing the size of its network \cite{bornschein2014, bornschein2016}. However, by increasing the number of neurons on a model's network, the size of the search space also grows. When any model is searching through the energy surface it can easily get stuck at a sub-optimal local minima \cite{hertz1991}, and we believe this is the main problem of the HM with a large hidden network. 

Our proposition to overcome this problem is to provide the algorithm with a heuristic for it to be more consistently led to optimal solutions.

Heuristics consist of ways to navigate the search space, that guide the algorithm to either find a better solution, find a solution faster, or both. They can be seen as generic rules that apply to a majority of the cases, allowing the agent to avoid exploring search paths that seem unpromising.

\subsection{Multi-level Data Representation and Human Image Perception}

One thing that might help humans understand what they see in a better and more structured way, is the ability to evaluate a given visual image at different scales. Many studies point to the fact that the human brain processes visual inquiries at different resolutions \cite{wilson1979, campbell1968}. This multi-level biological visual analysis could be one of the many keys that enable the human brain to capture the world it perceives in such a robust and accurate way despite the obvious extreme complexity of its neural network.

\begin{figure}
\includegraphics[width=\textwidth]{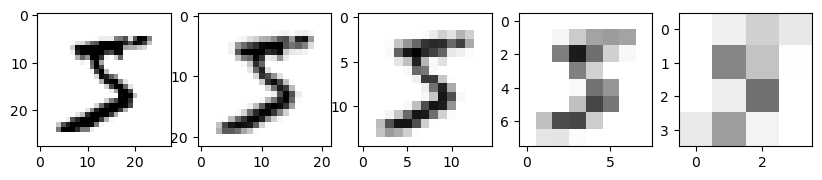}
\caption{Example of an Image Pyramid representation of a handwritten number 5 generated by continuously downsampling the original image on the left.} 
\label{fig:IP}
\end{figure}

A way to incorporate this multi-level perception into the HM is by using an Image Pyramid representation of the dataset \cite{rosenfeld1984}. The Image Pyramid is a simple way of having multi-level data representation that enables models to detect patterns on different scales. It consists of creating lower-level representations of the original images in a convolutional fashion, reducing an image by a factor each time, and creating a ``sequence of copies of an original image in which both sample density and resolution are decreased in regular steps" \cite{adelson1984}, like shown in Fig. \ref{fig:IP}. Introducing this data representation to the training of the network would be in accordance with the high biological plausibility that motivated the interest in the HM model and by doing so we hope to guide its learning, in a way that first detects high-level patterns, and then as we add details to the samples, it would learn more correlations on different scales, acting as a heuristic to overcome the exponential increase of the search space that inevitably comes with the increase of the number of hidden layers.

\subsection{Image Pyramid Heuristic for Helmholtz Machines}

\begin{figure}
\includegraphics[width=\textwidth]{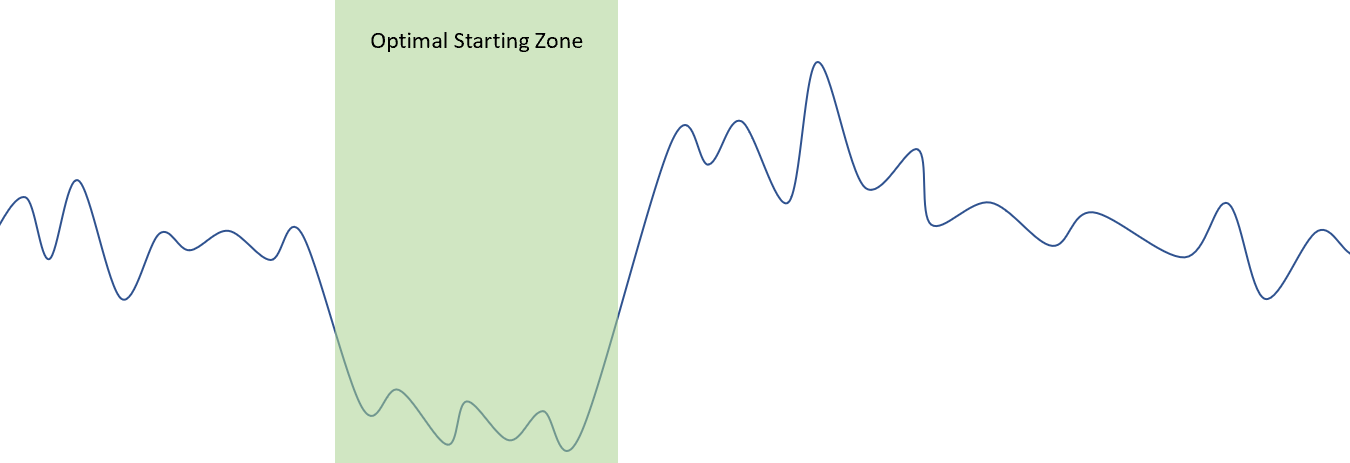}
\caption{Example of a two-dimensional energy landscape described by a blue curve. When traveling the energy surface with a non-stochastic gradient method, our model would move in a way similar to a sphere being dropped in the said landscape, moved by the force of gravity. We can understand that the starting configuration of our model, meaning, the starting position on the landscape, would have a major impact on the absolute value of the minima achieved. There is in this case an optimal starting zone that we highlighted in green, where if the initial configuration corresponds to a point in that zone, the minima reached would be generally better.} 
\label{fig:EL_OSZ}
\end{figure}

One way to guide our model's training is by configuring its initial position on the search space, to a zone where we believe the probability of finding a smaller local minima is higher like the one highlighted in Fig. \ref{fig:EL_OSZ}.

Weight initialization has been known to have a significant impact on the model's convergence state when training with deep neural networks \cite{sutskever2013, glorot2010}.
The idea of the heuristic we want to apply to the learning of the HM is to initialize the weights of the network so that the initial configuration contains queues of the image particularities at different scales.

We propose to create a network with multiple hidden layers, with increasing sizes from top to bottom where each layer must correspond to the size of a down-sampled image.

\begin{figure}
\includegraphics[width=\textwidth]{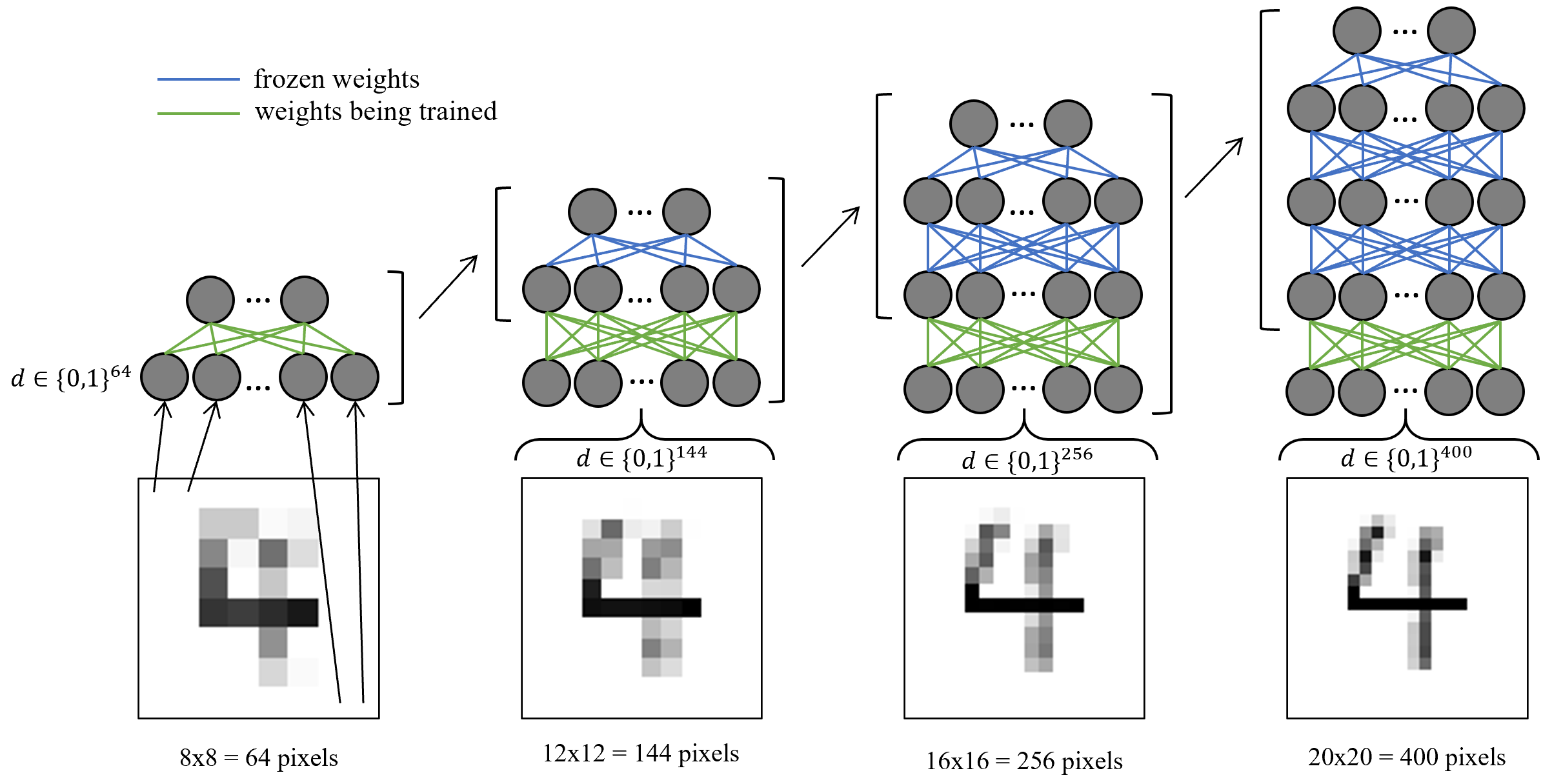}
\caption{Proposed Weight Initialization for a Helmholtz Machine. We use down-sampled images to train the smaller hidden layer and proceed to freeze the learned weights, then we up-sample the previously used images and train a newly added layer, then we freeze the new layer's weights repeating the process until we reach the original images' resolution. This way, information on all detail levels should be present in the initialization of the weights.} \label{fig:WIHM}
\end{figure}

Then we iteratively train the machines layer by layer, starting from a low-resolution sample of the original world's images \cite{wichert2008, wichert2015,  wichert2010}, and add additional layers while increasing the images' resolution as described in Fig. \ref{fig:WIHM}. The downsampled images would be equivalent to the idea of a blurred image where only the global details could be retrieved, and therefore, the first layer trained would in theory be able to recognize global features of the world's distribution. After learning a good distribution for a certain level of resolution, the model would then freeze the weights it learned for this layer, preventing it from losing its global perception when learning with more detailed data. Then we would increase the resolution and train an additional layer the same way we did with the previous one.

When we reach the last layer, we should have incorporated in our machine's weights the information of all resolution levels, and after it, we would conventionally train the HM, with the predetermined initial configuration.

\section{Results and Experiments}

In this section, we will propose and perform several experiments to confirm our previously stated hypothesis and test if our proposed heuristic provides significant advantages in the generative performance of the Helmholtz Machine.

We will first use the MNIST dataset of handwritten digits \cite{lecun1998} to train our models and perform our experiments. This dataset has a relatively small complexity but still allows us to compare results for different implementations in a permissive environment and to gather insights that could otherwise become imperceivable intricacies in more complex domains. Moreover, results on this dataset motivate future experiments on more complex datasets and act as a perfect stepping stone from conception to practical usage of any model.

After performing the proposed experiments on this dataset, we will test our heuristic on two other datasets, Fashion-MNIST \cite{xiao2017} and  CIFAR-10 \cite{krizhevsky2009}. Both of these datasets have higher complexity than the MNIST dataset of handwritten digits, with CIFAR-10 having the highest complexity of the three.

\subsection{Is the locality of Wake-Sleep a problem when training Deep Networks?}

One factor that may penalize the Helmholtz Machine's performance with deep architectures is the locality of the Wake-Sleep algorithm.

When adding hidden layers to our model, we are increasing the number of free parameters, so in theory, we would be increasing the network's potential to represent the world's data. However, we believe that the HM does not take full advantage of this augmentation in capacity, due to the fact the local updates present on the learning rule make the learning progressively harder to be propagated throughout consecutive hidden layers.

\subsubsection{Proposed Experiment:}

To test this hypothesis, we trained a Helmholtz Machine with a deep architecture, and used its recognition layers' activations as inputs for a simple Logistic Regression (LR) Model, to see how well the HM's hidden representations are able to linearly separate the problem space.
This approach takes advantage of the fact that our model is simultaneously training a recognition and a generative model. The quality of the generative model is related to the capability of our model to generate good lower-level explanations of the observed samples with its recognition network. So by testing our model's input representation at different steps of the recognition chain, we can see what hidden layers are responsible for identifying the majority of the learned features. 
For this experiment, we chose an architecture with 6 hidden layers of size 625 ($25\times25$), 484 ($22\times22$), 289 ($17\times17$), 196 ($14\times14$), 100 ($10\times10$), 16 ($4\times4$) starting from the input layer.

\subsubsection{Results:}

\begin{figure}
\includegraphics[width=\textwidth]{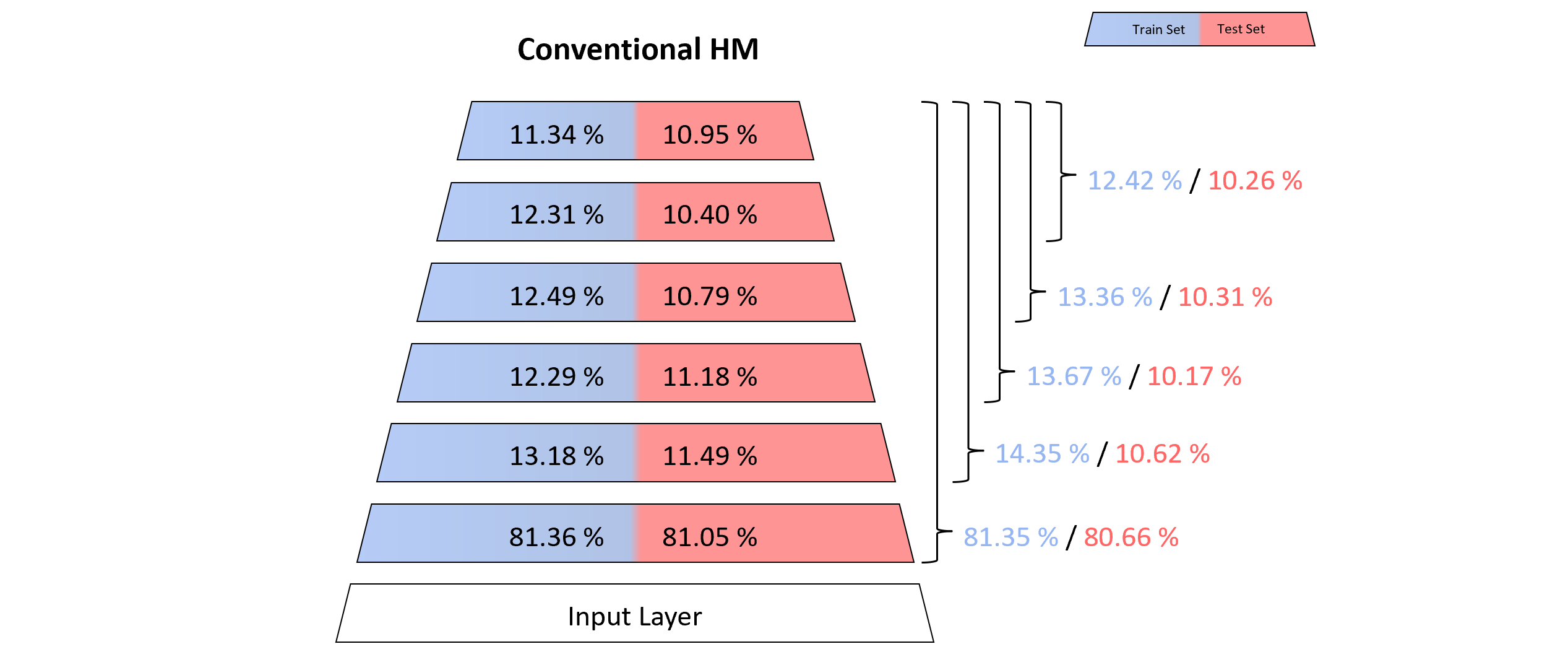}
\caption{Accuracy of the LRs trained with different subsets of Helmholtz Machine's Recognition Network's hidden representations of the MNIST samples, both in the train and test set. The values on each layer correspond to the usage of the neuron activations on that single layer, while the brackets correspond to the concatenation of activations on the layers they aggregate. We can see that most of the class separation is done in the first layers, whereas layers that are further away from the input layer bear almost no information about the world's distribution.} \label{fig:LR_Rand}
\end{figure}

From the results described in Fig. \ref{fig:LR_Rand} we can see that the majority of the separation of the problem space is done in the first layer. This suggests that with the local WS learning rule, as the size of the network increases, a large part of the information will not be propagated through the network, and will store most of the information regarding the learned features at the surface of the deep network, meaning that even tho we are adding more descriptive power to the model by increasing its depth, it is incapable of taking advantage from it.

\subsection{Does the Multi-level Data Representation solve this problem?}
\label{IP_LR}

\subsubsection{Proposed Experiment:}

We repeat the experiment proposed in the previous section, training a HM with the same architecture as the previously described one, but this time initialized with the proposed Image Pyramid method.

\subsubsection{Results:}

\begin{figure}
\includegraphics[width=\textwidth]{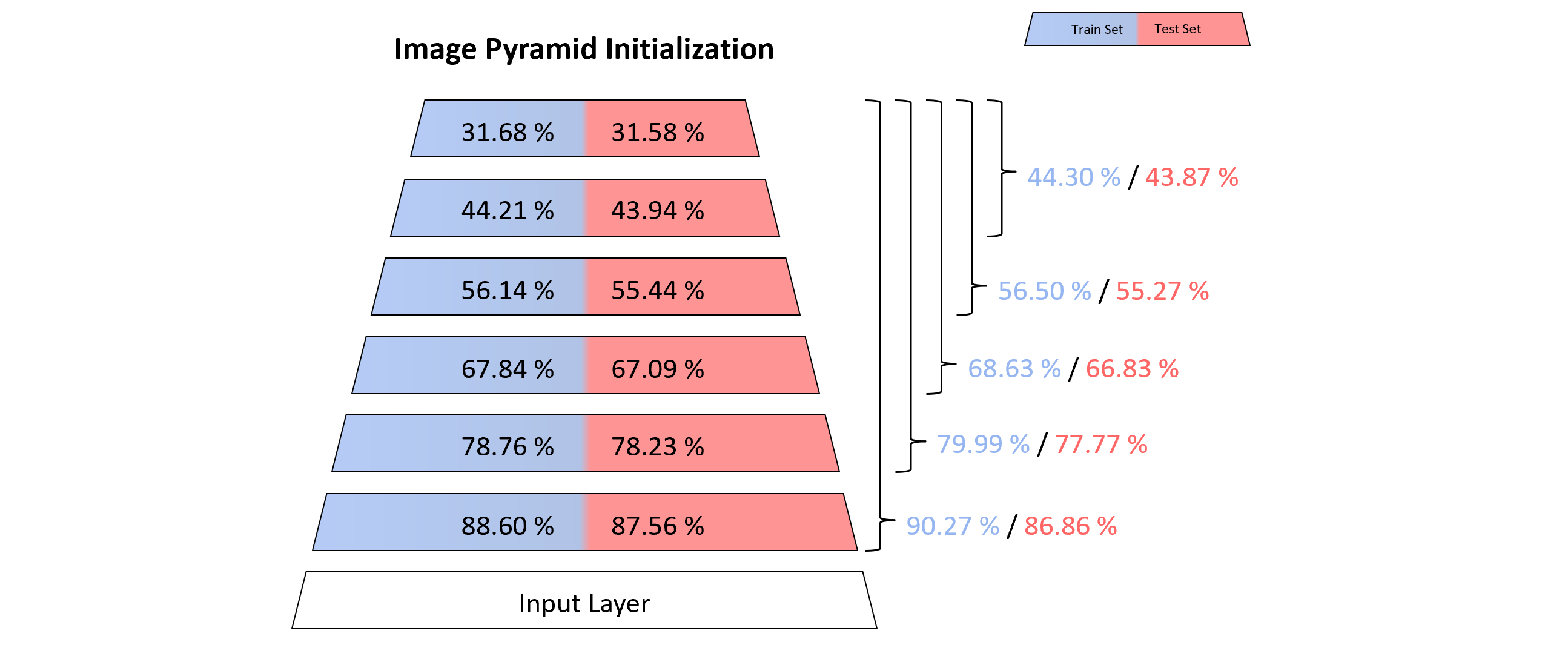}
\caption{Accuracy of the LRs trained with different subsets of Helmholtz Machine's Recognition Network's hidden representations of the MNIST samples, both in the train and test set. The values on each layer correspond to the usage of the neuron activations on that single layer, while the brackets correspond to the concatenation of activations on the layers they aggregate. Image Pyramid Initialization shows a progressive increase of the class separation capability as we get closer to the input layer, suggesting a better use of the network as a whole to define the main features of the samples.} \label{fig:LR_IP}
\end{figure}

In this experiment, the results presented in Fig. \ref{fig:LR_IP} show not only higher overall accuracy values, but more importantly, a smooth decrease of the layer's descriptive power as we reach higher layers, which is expected since the number of neurons on each layer is smaller as we go up the network.
These results are a good indicator that our heuristic provides an advantage for the recognition network's world representation, rendering it capable of fully using its deep hidden layers to store meaningful information.

\subsection{Does Multi-level Data Representation provide a generative advantage?}

In the previous section, we focused solely on the evaluation of the Recognition Network. We presented evidence for our claim that the Image Pyramid Initialization allows for better usage of the capacity of the deep network, and hypothesized that a better Recognition Model would also be translated into a better Generative one.

Consequentially, we should be able to see a similar improvement when testing the Generative Network and prove that when using Image Pyramid Initialization, we take full advantage of the network’s increase in size. 

\subsubsection{Proposed Experiment:} 
\begin{figure}
\includegraphics[width=\textwidth]{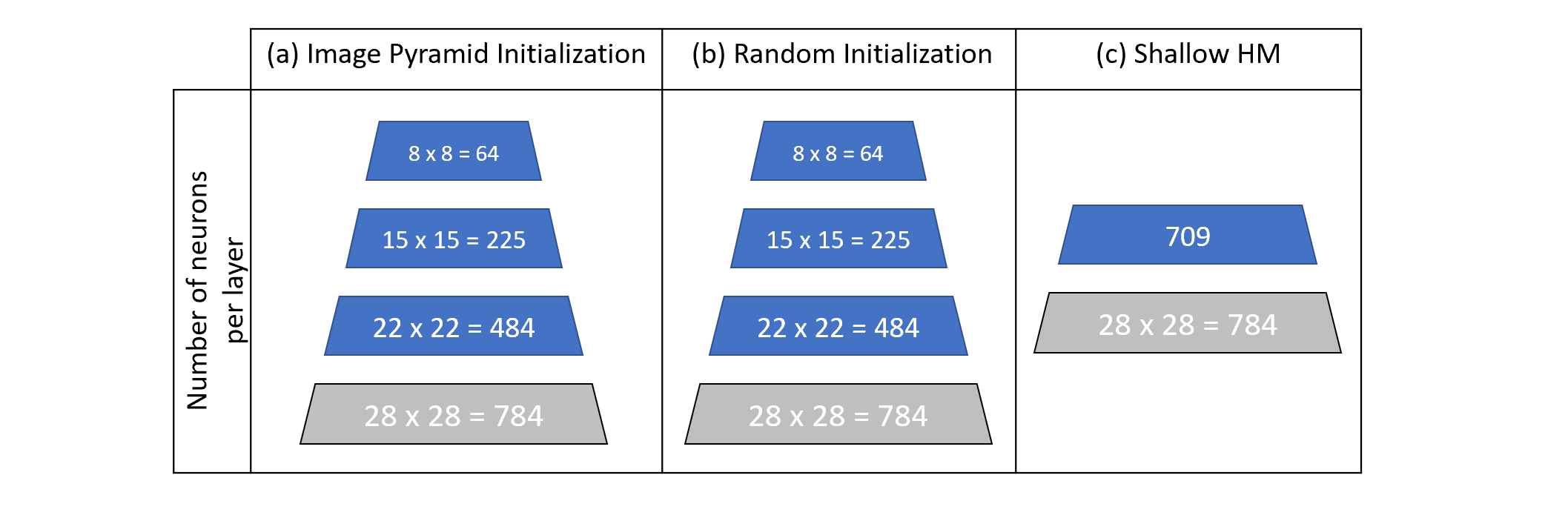}
\caption{Proposed architecture for 3 different Helmholtz Machines. Machines (a) and (b) have three hidden layers and an equal architecture with a number of free weights $\rho \approx 64 + 64 \times 225 + 225 \times 484 + 484\times784=502820$, while machine (a) consists of a single hidden layer machine with $\rho \approx 709 + 709\times784=556565$. Thus, the descriptive power of the machines should follow the same order as the number of free parameters $\rho_{(a)} = \rho_{(b)}  < \rho_{(c)}$. Machines (b), and (c) are initialized with random values, whereas machine (a) is initialized using our proposed multi-level representation method.} \label{fig:Gen_exp}
\end{figure}

To test our hypothesis, we will define an architecture for a Neural Network and create two different machines with that same architecture. One of the machines will use the Image Pyramid Initialization (Fig.  \ref{fig:Gen_exp} a), and the other will use a classic Random Initialization (Fig.  \ref{fig:Gen_exp} b).
After, we proceed to train them with a small train set of size $N$ (e.g. 2 samples) and see if the network is able to generate it back. To do this, after the machine has been trained, we generate a large number of samples $G$, and find the euclidean distance from a given sample to the train set. Then, we choose the minimum distance observed, and claim that the generated sample corresponds to that particular train set image. We keep the smallest distance observed and the correspondent train set sample and repeat the same process for all generated samples. We end up with an array of closest distances, and an array of the correspondent train set samples. With the array of distances, we simply calculate the mean, and with the correspondence array, we first create an array with the size of the Train Set where each index corresponds to the representation fraction of the same index train sample in those $G$ generations, creating a density vector (e.g. following the previous supposition that we only have 2 samples, $x_0$  and $x_1$, if the model generated 6 samples closer to $x_0$ and 4 samples closer to $x_1$, the corresponding density vector would be $[0.6, 0.4]$). From that density vector, we take two different measures, the first one being the Entropy, and the second the Number of Unrepresented Samples. The Entropy will be close to one if the machine generates the same number of samples for each Train Set image, and closer to zero as the model starts to replicate some of the world's images more frequently than others, so in general this measure relates to how well the machine is capturing the real world's distribution. The Number of Unrepresented Samples shows how many samples the machine has ``forgotten", meaning that it was unable to closely replicate a sample present during its training, despite having generated a large pool of samples. If this measure is 0, the machine was able to remember all learned samples, but as the number of samples in the world increases, the machine will inevitably become unable to represent some of them. With this measure, we can see the breaking point regarding the world's number of samples at which the model becomes unable to remember all samples seen and can be used as a comparison measure between different models.

In addition to our initial claim, we also believe that using a deep network with Image Pyramid Initialization with a certain number of free parameters has a generative advantage when compared to a shallow one-layer network with a higher number of free parameters given the compositional properties that multi-layer networks allow for. Therefore we include in the collection of machines an additional large shallow HM with a hidden layer size equal to the number of neurons of the two bigger hidden layers of the deep architecture (Fig.  \ref{fig:Gen_exp} c), assuring it has a higher number of free parameters.

\subsubsection{Results:}

\begin{figure}
\includegraphics[width=\textwidth]{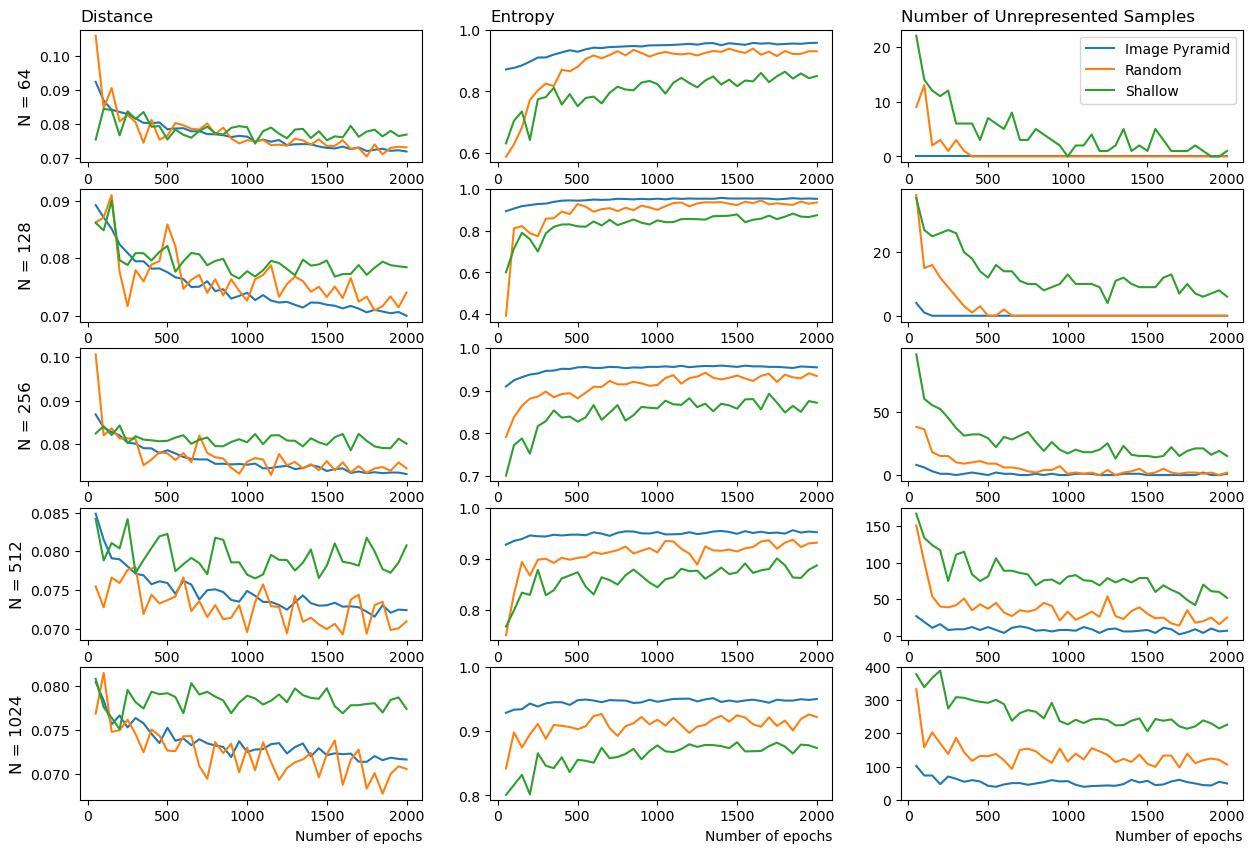}
\caption{Measures for the three proposed machines regarding Mean Distance, Entropy, and Number of Unrepresented Samples on the first, second and third column respectively, with the number of train set samples $N$ increasing on each row. The multi-level based approach showed better results on both Entropy and Number of Unrepresented Samples for all $N$ values, while the shallow network performed worse on all measures for all experiments.} \label{fig:Train_gen}
\end{figure}

The results in Fig. \ref{fig:Train_gen} are very promising and suggest that our initial intuition was true. Regarding the Number of Unrepresented Samples, we can see that up to $N=64$ all machines can fully represent the dataset, from 128 to 256 the shallow machine starts to be unable to represent certain samples, while the deeper architectures can still fully represent them. From 512 to 1024 we can start to see a difference in the performance of the Random to the Multi-level based initialization, giving an edge to the latter one. The Image Pyramid machine was able to have a higher entropy for all $N$ values followed by the deep machine with Random initialization and lastly the shallow one, suggesting our approach is able to gather a better generalization of the world's distribution. The mean distance is similar in both deep architectures and consistently better than the shallow machine for all $N$ values.

The poor performance of the shallow network with a higher number of free parameters indicates that the compositional properties of multi-layer can help a generative model capture the real world's distribution better. 

The edge that machine (a) had over machine (b) when representing its training dataset suggests that using multi-level data representation provides advantages to the generative capabilities.

\subsection{Can we quantify the generative advantage of Multi-level based Initialization?}
\label{gen_adv}

From the previous experiment, we saw that the HM was able to replicate more samples on a given dataset when using our proposed initialization, which is a good indicator that the machine can understand the world's general distribution. However, we think mimicking a train set is not the goal of a generative model.

Testing a Generative model's performance is not a trivial task \cite{alaa2022}, and up to this date, there is no perfect definition of what can be considered a good generation, since different problems focus on different generative goals. Therefore, there is also no evaluation method devoid of criticism \cite{theis2015}.
With this in mind, we decided to enumerate what we thought were the desired attributes our machine's generation required, in this particular experiment, with regards to handwritten digits' image generation.
The most important attribute was the quality of the generated samples, more specifically, how similar the generated patterns were to real handwritten digits.
The second attribute was diversity in the generated samples. 
And lastly, the propensity of generating new patterns. 
This last attribute might seem counter-intuitive, but the idea is that if our model produces completely different images of a digit that it did not see in the train set, but follow its digits general rules (eg. for number 8 two circles attached vertically), then our model effectively learned the core defining features of a digit.

\subsubsection{Proposed Experiment:}
We decided to take a common approach of tweaking the original generative model so that it can be used as a classification one. With classification, the model's performance becomes much easier to quantify, since we can get concrete measures such as error and accuracy. The fact that our machine learns with unlabelled data makes it hard for our model to be used to classify digits, so we decided to create 10 different machines, one for each digit. Hopefully, each machine's generation corresponds solely to good representations of its designated digit (what we call good quality) and produce a wide variety of that digit's possible representation (what we call variety).
Then, we generate a fixed number of samples from each of the ten machines, and we end up with an entirely new generated dataset by combining all samples.
After we create the new dataset, we can associate labels to the generated patterns, since each sample is associated with a certain digit's machine. Now, we can use a simple classification model like a K-Nearest Neighbor (KNN) trained with the generated dataset to classify the test set.
We decided to use a KNN with $k=1$ because of its simplicity. We believe it is a good choice because the score of the KNN's performance is purely related to the quality of the dataset, and our ultimate goal is not to create the best possible classifier but to test the quality of the generated dataset.
If our machines can produce a wide range of variations of its designated digit, we should end up with a dataset that is able to produce the possible digit variations existent on the test set, and thus allow the KNN to have better accuracy during the test phase.
We believe this evaluation method favors models that have the three requirements previously enumerated, but we can see scenarios where solutions that do not meet all the requirements still perform significantly well. For example, a machine generates the pixel distribution of a certain digit's class (similar to performing a mean of all the digit's samples), despite not having any variability, the KNN would still in most cases associate a test sample to the correct machine.
To ensure variety amongst the generated samples we decided to measure the average euclidean distance to the mean of the newly generated dataset (ADM) similar to variance in a standard deviation, and to ensure that the generated instances are different from the training dataset we calculated another indicator called Novelty, that is obtained through the sum of the smallest distance of each Train Set sample to the generated ones.

Lastly, we performed said experiment on three different initialization methods, Zero Initialization that assigns all initial weight values to 0, Random Initialization that uses random values across a standard deviation centered on 0, and the Image Pyramid Initialization proposed.

\subsubsection{Overall results:}
\label{overall_res}
\begin{figure}
\includegraphics[width=\textwidth]{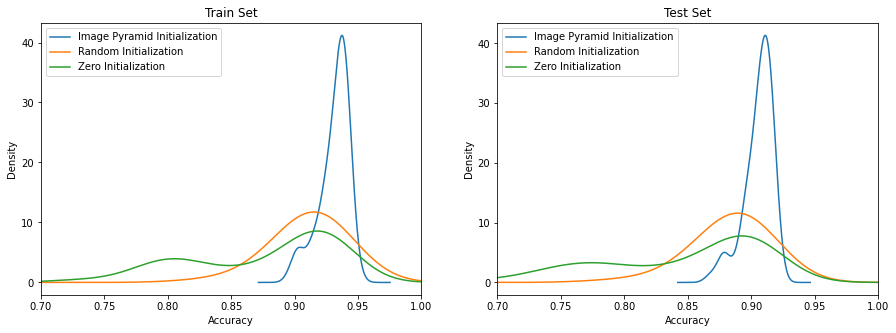}
\caption{Density function for the train and test set accuracies of 70 randomly generated architectures trained with Random, Zero, and Image Pyramid Weight Initialization. Multi-level Representation Initialization shows not only a higher overall accuracy but also a lower variety, showing not only better performance but also more robustness.} \label{fig:Acc_overall}
\end{figure}

The accuracy obtained with different network architectures and with different weight initialization methods described in Fig. \ref{fig:Acc_overall}, shows a clear advantage for initialization based on multi-level data representations, having not only a better average score but also smaller variance.
We believe these results indicate that the Image Pyramid Initialization guides the networks' learning in a more robust way, by starting in an area of the energy surface where the local minima reached are generally better.

\begin{figure}
\includegraphics[width=\textwidth]{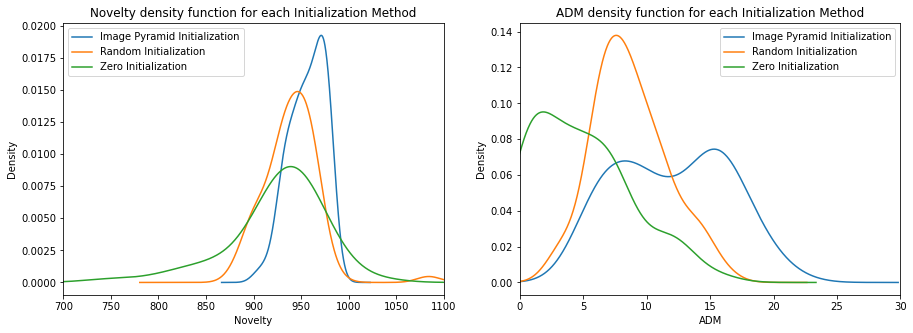}
\caption{Density function of two different variability measures. Novelty relates to the difference between the generated data samples to the training samples, whereas ADM relates to variability among the generated samples. Our proposed Initialization method generated not only instances that were more different amongst themselves, but also less similar to the observed ones.} \label{fig:Var_overall}
\end{figure}

When looking at the variability measures for the same run of experiences in Fig. \ref{fig:Var_overall}, the results show that our proposed method is able to generate samples more different from each other, and also samples less similar to the training set. This latter factor is a very promising indicator when combined with the previous observation that the accuracy performance also increased. This could indicate that the model not only was able to produce new images, but those images are viable candidates for existing in the real world, possibly very similar to the unseen samples in the Test Set.

\subsubsection{Results regarding dimensionality:}

\begin{figure}
\includegraphics[width=\textwidth]{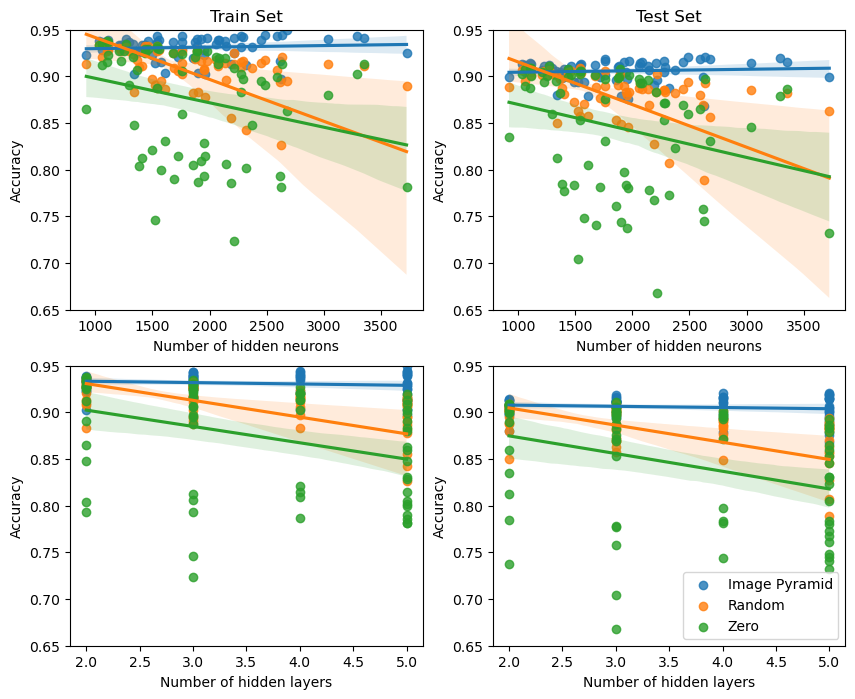}
\caption{Scatter Plot with a Linear Regression Fit of the accuracy for Random, Zero, and Image Pyramid Initialization both on the Train and Test Set, plotted with regards to the number of neurons on the bottom axis on the first row, and to the number of hidden layers on the second one. Both Random and Zero Initialization show a progressive accuracy drop with the increase of the network's size, while the Image Pyramid's accuracy remains unchanged.} \label{fig:Acc_size}
\end{figure}

When we plot the accuracy with regards to the complexity of the model (either by the number of neurons or the number of hidden layers) in Fig. \ref{fig:Acc_size} we can see a clear decrease in the accuracy for the random and zero initialization, whereas the multi-level representation initialization remains consistent with the complexity increase. We believe these results show a clear inability for normal HM to perform well on deeper networks, which can easily be fixed by adding multi-level image representations to its learning. However, we expected the Image Pyramid Initialization to see an increase in accuracy with the increase of the number of free parameters, which did not happen on a meaningful scale. One of the reasons for this occurrence might be related to the small complexity of the dataset. We initially hypothesized that more free parameters were necessary for describing more complex worlds, however, if the world's complexity is small, there is no need for more free parameters. The MNIST dataset of handwritten digits is known to have relatively low complexity, and thus, as we surpass the required number of free parameters, we should not see any meaningful increases in the model's performance.

\subsection{Can a Helmholtz Machine transcend the Train Set?}
\label{KNNSurpass}

Encourage by the results mentioned in section \ref{overall_res}, where we stated that the Helmholtz Machine using the Multi-level heuristic was able to produce a wider variety and more creative samples, we decided to test if a HM was able to generate a dataset that could surpass the KNN performance of its original train set. We believe that after learning all digits, if a human spent an enormous amount of time generating labeled digits variations, eventually, a simple KNN using that generated data could almost perfectly classify the unseen MNIST test set. Likewise, if a HM produced enough samples it could cover a wider range of possible test set instances.

\subsubsection{Proposed Experiment:}
To test this hypothesis, we decided to train three 10-machine's models with the same architecture and each initialization method used previously. The architecture chosen was a 2 hidden layer network with layer sizes  400 ($20\times20$) and 100 ($10\times10$), while still having a visible layer of size 784 ($28\times28$). We defined a Train Set of 10000 samples from the MNIST dataset and tested the performance of the chosen Train Set on the Test Set defining a threshold for our model to try to surpass. After we trained the machines with the chosen Train Set, we decided to calculate the accuracy of a KNN using datasets generated by the 10-machine model, and see what would happen with the increase of the size of the generated data, previously denoted as $G$.

\subsubsection{Results:}

\begin{figure}
\includegraphics[width=\textwidth]{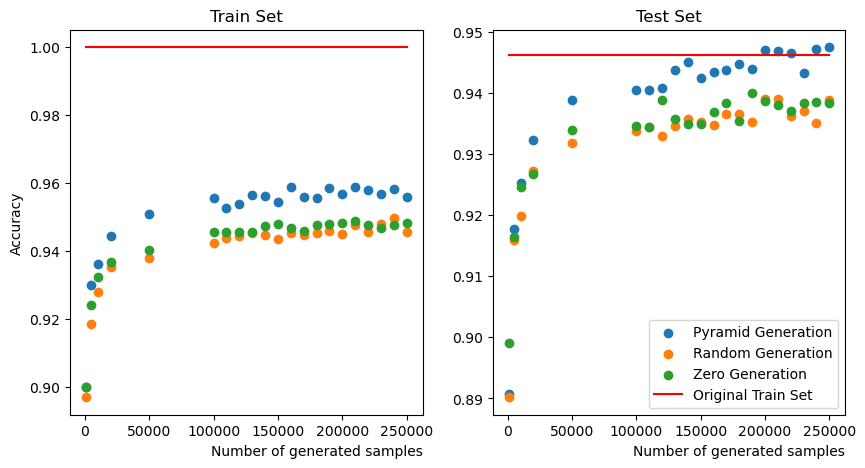}
\caption{Scatter plot of the accuracy in the Train and the Test Set of a 1NN classifier using datasets generated by the HM models with three different initialization algorithms, plotted with regards to its number of samples. The blue horizontal line corresponds to the accuracy of a 1NN classifier using the same Train set the HM models used for training. We can see that only the model using the Image Pyramid initialization was able to surpass that horizontal line.} \label{fig:KNN}
\end{figure}

From the results in Fig. \ref{fig:KNN} we can see that only the model trained with our multi-level heuristic was able to reach the same accuracy as the original Train Set, even surpassing its performance by a small margin, which strongly indicates that the model can generate samples more similar to the test set than the ones existing on the Train Set.

\subsection{Is the Generative Advantage still present on more complex Datasets?}

To conclude our experiments we believe it is important to understand if our heuristic still provides advantages in different and more complex domains. We decided to use the Fashion-MNIST and the CIFAR-10 datasets for this purpose. In the CIFAR-10, there are three color channels, and although it is possible to create a HM architecture to address this, we believe that changing the architecture of the model for this particular experiment would ravel the comparison to the other datasets, so we decided to change the RGB triplet of the images in CIFAR-10 to a grayscale, allowing us to have similar HM architectures for all domains.

\subsubsection{Proposed Experiment:} 
We decided to perform a similar experiment as we did in section \ref{gen_adv} using the 10-machine model to generate a dataset followed by a KNN using the generated dataset to classify the Test Set. Since the classification task is generally harder as we increase the complexity of the world, the overall accuracy values obtained will decrease as the complexity of the world grows. We believe the fairest comparison measure would not be the total KNN accuracy, but the accuracy improvement using the generated dataset, compared to the accuracy obtained using the Train Set. So the proposed metric to compare the results would be the Accuracy Improvement Factor, obtained by dividing the accuracy of the KNN using the generated dataset by the accuracy of the KNN using the original Train Set.

\subsubsection{Results:} 

\begin{figure}
\includegraphics[width=\textwidth]{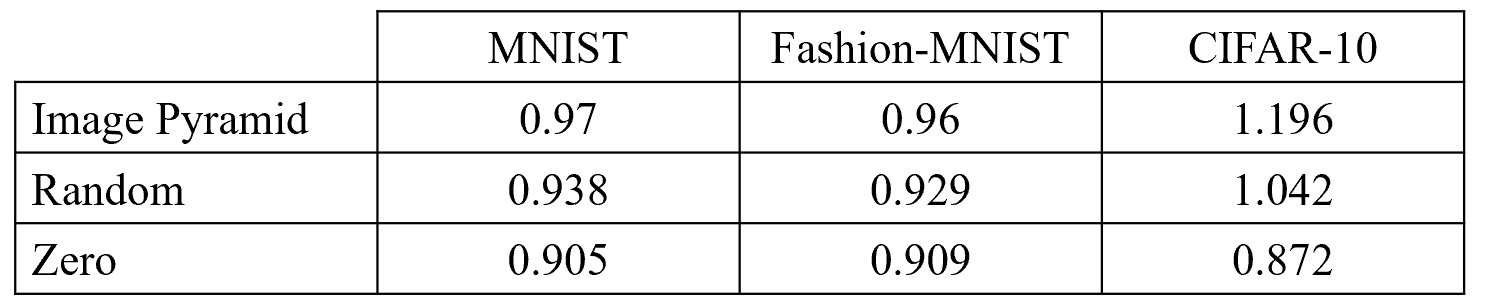}
\caption{Accuracy Improvement Factor on the Test Set for the three Initialization methods in each row, on different Datasets in each column. There is a clear advantage when using the Multi-level approach compared to the more classical implementations at all levels of domain complexity.} \label{fig:AIF}
\end{figure}

From the results in Fig. \ref{fig:AIF}, we can see that across all three datasets, each one with a different complexity, the performance of the model that is using the proposed heuristic was higher, which suggests that the advantages observed in the simpler domain of the MNIST dataset of handwritten digits are also present when the complexity of the world increases.

\section{Conclusions and Future Work}

Despite the undeniable success of global Gradient-based algorithms, to understand the underlying mechanisms of biological intelligence it is necessary to develop models whose implementation could be plausible in a biological neural network. 
We believe a great candidate for said implementation is the Helmholtz Machine, due to the biological inspiration and locality of its training algorithm. 

We hypothesize that, unlike gradient algorithms such as Back-propagation, this local learning algorithm does not perform well under deep network architectures, making it difficult to take advantage of the composition properties that multi-layer networks provide.

To test this hypothesis, we trained a HM on the MNIST dataset of Handwritten digits and used a linear classifier trained with a subset of the hidden representations of the HM's recognition model. The results were compliant with our hypothesis and showed that most of the separation was concentrated in the first layers, whereas deeper smaller layers had almost no relevant information regarding the problem space.

To avoid this limitation of the learning algorithm, we came up with a heuristic for the initialization of the machine's weight vectors by using a multi-level data representation based on the idea that humans process visual inquiries at different resolution levels.
By iteratively training the smaller layers with downsampled images of the real dataset, and increasing the resolution as we increase the size of the new layers, we believe that the information regarding high abstract levels of representation is rendered into each layer, obtaining the core defining features of the correspondent resolution. 

Using our proposed solution, and repeating the previous experiment we saw that the separation of the problem space was done uniformly throughout the deep network, suggesting a better usage of the network as a whole in the recognition model of the HM.

We then saw that the model's improvement was also imminent in the generative model, showing that our proposed solution was able to take advantage of the compositional properties that multi-layer networks allow for, being able to replicate a more complex world than a one-layer network with a higher number of free parameters and an equal architecture trained conventionally.

We performed further experiments to test the generative advantages of adding multi-level information to the training of the neural network, by using the model's generated images to train a simple classifier and concluded that the machine's generated samples had not only better quality, since the classifier performed generally better when using our approach, but were also more diverse and creative when compared to the classic implementation.

From the latter experience, we were able to find further evidence that supports our claim that the WS algorithm is not fit for Deep Networks, showing a progressive decrease in performance as the network's depth increases when using classical implementation. The same occurrence was not imminent when using our proposed initialization step.

Encouraged by the creativity measures of our model, we decided to see if it was able to produce a dataset for training a classifier that would outperform an equal classifier that learned on the HM's initial training set. We saw that a such thing was possible when generating a large number of data samples but only with our proposed multi-level heuristic.

Finally, we detected that the Multi-level heuristic still provides generative advantages in more complex domains, by performing similar experiments on different datasets.

The results obtained were very promising and showed the innate potential of the Helmholtz Machine's generative model's capabilities. Moreover, we believe the general heuristic idea can be applied to other local learning algorithms training on two-dimensional data with similar success.

\section{Declarations}

This work was supported by national funds through Funda\c{c}\~{a}o para a Ci\^{e}ncia e Tecnologia (FCT) with reference UIDB/50021/2020 and through a doctoral grant SFRH/BD/144560/2019 awarded to the second author. The funders had no role in study design, data collection and analysis, decision to publish, or preparation of the manuscript. The authors declare no conflicts of interest. Code and data for all the experiments can be obtained by email request to the first author.

%
%
%
%

\printbibliography
\newpage

\appendix

\section{The Wake-Sleep Algorithm}
\label{WS_details}

As we said previously, during the HM model's learning, the WS trains both a Recognition $(R)$ and a Generative $(G)$ network in two different phases, the Wake-phase and the Sleep-phase.

In the Wake-phase, it updates the generative weights to try to minimize the Variational Free Energy, which given a pattern $\mathbf{d}$ and hidden-layer activations $\mathbf{H}$ is described by:

\begin{equation}
F^R_G(\mathbf{d}) = F_G(\mathbf{d}) + KL[p_R(\mathbf{H}|\mathbf{d}), p_G(\mathbf{H}|\mathbf{d})].
\end{equation}

And in the Sleep-phase it updates the recognition network trying to minimize:

\begin{equation}
\widetilde{F}_{G}^{R}(\mathbf{d})=F_{G}(\mathbf{d})+\mathrm{KL}\left[p_{G}(\mathbf{H} \mid \mathbf{d}), p_{R}(\mathbf{H} \mid \mathbf{d})\right].
\end{equation}

Where 

\begin{equation}F_G(\mathbf{d}) = - log \, p_G(\mathbf{d})\end{equation}

${F}_{G}(\mathbf{d})$ is a function called Free Energy that is inverse to the probability of $\mathbf{d}$, so will be high when this pattern is unlikely, and low when it is likely. Thus, this is also referred to as the ``surprise of $\mathbf{d}$", denoting how surprised the model is when observing the occurrence of pattern $\mathbf{d}$.

We can see that the Variational Free Energy is the Free Energy of a pattern plus a KL divergence of the Recognition and Generative probability distributions. From this we can understand that the Variational Free Energy is an upper bound of the Free Energy since the KL divergence is always positive, so minimizing the Variational Free Energy also minimizes the Free Energy. Additionally, minimizing the KL divergence means that some part of the network's training involves trying to make the Recognition and Generative networks congruent, meaning approximate inverses of each other.

The Wake-Sleep algorithm consists of multiple Wake and Sleep phases that continuously minimize the Variational Free Energy through a local gradient-based rule.

For a better understanding of the network's definition and the learning algorithm, we suggest looking at Kevin G. Kirby's tutorial \cite{kirby2006}, which provides an in-depth, clear, and intuitive explanation of the HM model.

\end{document}